\documentclass[letterpaper, 10 pt, conference]{ieeeconf}  
\usepackage{booktabs}
\usepackage{multirow}
 \usepackage{amsmath} 
 \usepackage{graphicx}

\IEEEoverridecommandlockouts                              

\title{\LARGE \bf
Enhancing Surgical Robots with Embodied Intelligence for \\Autonomous Ultrasound Scanning
}

\author{Huan Xu$^{1\dag}$, Jinlin Wu$^{1,2\dag}$, Guanglin Cao$^{1,2}$, Zhen Lei$^{1,2}$, Zhen Chen$^{1*}$, and Hongbin Liu$^{1,2}$
\thanks{$\dag$Equal contribution, $*$ Corresponding author.}
\thanks{This work is supported in part by InnoHK program.}
\thanks{$^{1}$Centre for Artificial Intelligence and Robotics, Hong Kong Institute of Science \& Innovation,
Chinese Academy of Sciences.}
\thanks{$^{2}$Institute of Automation, Chinese Academy of Sciences.}
\thanks{ {Email: \{jinlin.wu, zhen.chen\}@cair-cas.org.hk}}}

\begin{document}

\maketitle
\thispagestyle{empty}
\pagestyle{empty}

\begin{abstract}

Ultrasound robots are increasingly used in medical diagnostics and early disease screening. However, current ultrasound robots lack the intelligence to understand human intentions and instructions, hindering autonomous ultrasound scanning. To solve this problem, we propose a novel Ultrasound Embodied Intelligence system that equips ultrasound robots with the large language model (LLM) and domain knowledge, thereby improving the efficiency of ultrasound robots. Specifically, we first design an ultrasound operation knowledge database to add expertise in ultrasound scanning to the LLM, enabling the LLM to perform precise motion planning. Furthermore, we devise a dynamic ultrasound scanning strategy based on a \textit{think-observe-execute} prompt engineering, allowing LLMs to dynamically adjust motion planning strategies during the scanning procedures. Extensive experiments demonstrate that our system significantly improves ultrasound scan efficiency and quality from verbal commands. This advancement in autonomous medical scanning technology contributes to non-invasive diagnostics and streamlined medical workflows.

\end{abstract}
\section{INTRODUCTION}
Ultrasonography (US) is crucial for non-invasive diagnostics and early detection across medical fields \cite{chan2011basics,shung2011diagnostic}, enhancing patient care and outcomes \cite{Mayo2019ThoracicUA,Robba2019BrainUM,Moore2011PointofcareU}. In particular, it is pivotal in diagnosing fetal abnormalities \cite{sonek2007first}, gallbladder stones \cite{cooperberg1980real}, and cardiovascular diseases \cite{nezu2020usefulness}, improving early diagnosis and management \cite{kasoju2023digital}. To integrate robotics with the US for better scanning efficiency and quality, extensive studies currently focus on autonomous scan path generation \cite{Merouche2016ARU,Jiang2021MotionAwareR3,Jiang2020AutonomousRS}. Despite these advances, dynamic adjustments of US robotics remain difficult, especially the instruction understanding and dynamic execution.

We propose an ultrasound embodied intelligence system merging US robots with large language models (LLMs) for improved clinical performance. Our system leverages LLMs for understanding doctors' intentions and enhancing motion planning, enriching LLMs with US-specific knowledge, APIs, and robot manuals for reliable workflows and error mitigation. A specialized embedding model aligns robot actions with doctors' intentions, enhancing clinical demand fulfillment and offering precise, efficient solutions. Additionally, a dynamic execution mechanism inspired by ReAct \cite{yao2022react} allows verbal command interpretation into precise scanning paths, reducing manual adjustment needs. This thought-action-observation cycle engages with robot APIs for seamless command execution, showcasing LLMs' potential to enhance robotic precision and autonomy in healthcare.
\begin{figure}[t]
    \centering
    \includegraphics[width=0.99\linewidth]{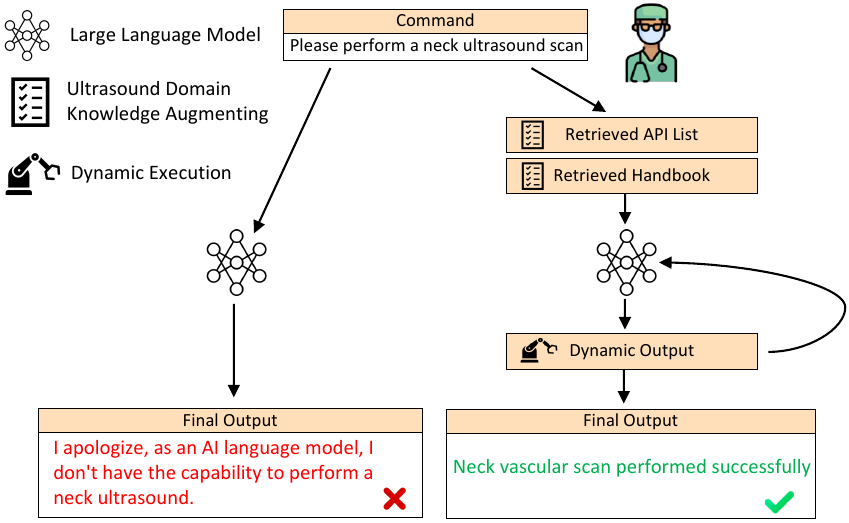}
    \caption{Our Embodied Intelligence system interprets and executes medical procedures through verbal commands. This system has three components: a large language model for command interpretation, the Ultrasound Domain Knowledge Augmenting technique for contextual understanding, and Robot Dynamic Execution for converting instructions into robotic actions.}
    \label{fig:system}
\end{figure}

\section{METHOD}
For a given ultrasound scan task, the scanning of the ultrasound robot can be formulated as follows:
\begin{equation}
     C_n = \prod_{i=1}^{n} R_i(C_{i-1}|A, U, D)),
    \label{eq:for}
\end{equation}
where
\begin{itemize}
    \item $D$ represents the Doctor's Instructions.
    \item $U$ denotes the Ultrasound Knowledge Augmenting.
    \item $A$ is the Assemble Ultrasound Assistant Prompt with retrieved APIs and retrieved robot handbook.
    \item $R_i$ signifies the $i^{th}$ iteration of interaction with robots through Robot Dynamic Execution.
    \item $C_n$ is the Final Execution Results, after $n$ iterations of dynamic execution.
\end{itemize}

\subsection{Ultrasound Domain Knowledge Augmenting}
\noindent \textbf{Domain Knowledge Search.}
To augment ultrasound scanning knowledge, we incorporate a similarity search algorithm that leverages cosine similarity to connect user queries with a database of ultrasound domain knowledge. This process involves converting queries and database entries into vectors in a d-dimensional space, aiming to find the highest cosine similarity match, as follows:
\begin{equation}
\mathcal{S}(\mathbf{A}, \mathbf{B}) = \frac{\mathbf{A} \cdot \mathbf{B}}{\|\mathbf{A}\| \|\mathbf{B}\|}.
\label{eq:cos}
\end{equation}

\noindent \textbf{Ultrasound APIs Retrieval.}
To streamline the selection of ultrasound APIs by LLMs, we refine the Ultrasound APIs Retrieval (UAR) method. This method relies on a dataset where each APIs is paired with a narrative describing its use context, improving tool selection for ultrasound scanning. The dataset is structured as $\mathcal{D} = \{(T_1, U_1), (T_2, U_2), \dots, (T_n, U_n)\}$, where T represents tools and U represents usage of tools, facilitating accurate tool identification based on scenario-specific requirements. 

\noindent \textbf{Robotic Handbook Retrieval.}
At the same time, how LLMs discern these API calls' correct order and logical sequence remains a huge hurdle. We present the Robotic Handbook Retrieval (RHR) method to address this issue. This approach enriches LLMs's context with a procedural knowledge base accessible through vectorized input queries. The core of this method is a similarity search. We systematically paired instructions with the handbook to guarantee the identification of relevant instructions during the similarity search, followed by the extraction of corresponding handbooks.

\subsection{Ultrasound Assistant Prompt}
Facing the challenge of commands lacking context, we enhance model comprehension and intent accuracy through structured prompts and added context. This approach ensures commands are interpreted precisely, aligning results with user expectations. Additionally, prompts are integrated with an execution session, allowing for specific output structures to trigger various APIs. 

\subsection{Robot Dynamic Execution}
Building upon the inspiration drawn from the ReAct framework mentioned in the introduction, we introduce a dynamic execution mechanism for robotic systems. This mechanism operates through a cyclical process comprising three main steps: Observation, Thought, and Action. This operational cycle aims to minimize errors and optimize task execution by continuously adapting to real-time feedback. 
\section{Experiments}
\subsection{Experimental Setup}
\noindent \textbf{Models Configuration.} We evaluated our system enhancements using the GPT4-Turbo model \cite{openai2023gpt4}, with settings at \textit{Temperature = 0.7} and \textit{Top P = 0.95}. The bge-large-en-v1.5 \cite{bge_embedding} model, adjusted for our domain, and FAISS \cite{douze2024faiss} were used for efficient vector operations. We compare this model performance against the original in Table \ref{tab:embed}.

\begin{table}[!h]
\centering  
\caption{Comparison of model training results for domain adaptation}  
\begin{tabular}{ccccc}  
\toprule  
Module & Model & Recall@1 & Recall@3 & Recall@10 \\  
\midrule  
\multirow{2}{*}{UAR} & bge-large & 0.82 & 0.94 & 0.97 \\  
 & Ours & \textbf{0.86} & \textbf{0.96} & \textbf{0.99} \\  
\midrule 
\multirow{2}{*}{RHR} & bge-large & 0.76 & 0.95 & 0.97 \\  
 & Ours & \textbf{0.88} & \textbf{0.97} & \textbf{0.98} \\  
\bottomrule  
\end{tabular}  
\label{tab:embed}  
\end{table}

\noindent \textbf{Datasets and Preprocessing.} Our synthetic dataset includes 522 Robotic Handbook instances and 622 Ultrasound API instances, reflecting the complexity of ultrasound scans and API usage. This dataset also trained our embedding model.

\noindent \textbf{Experimental Framework and Metrics.} The framework assesses the augmentations' impact through 20 repetitions of each experiment, comparing model performances and testing on various human body parts. 
\subsection{Results and Analysis}

We conducted ablation studies and model performance analyses, with initial findings in Table \ref{tab:combined_study}, where FS stands for the success rate of the first step attempting and OV stands for the success rate of all execution steps.

\noindent \textbf{Effectiveness of Modules.} Results show foundational LLMs improve stepwise with each module. Initially, LLMs struggle with API calls without specific knowledge. Introducing the UAR module improved success to 35\%, indicating basic API knowledge aids in task initiation. The RHR module further boosts performance, underscoring structured guidance's importance in API selection and task-specific improvements.

\noindent \textbf{Effectiveness of Different LLMs.} Evaluating various LLMs augmented with domain knowledge showed differing success rates in API call execution and task completion. The Mixtral-8x7B-Instruct-v0.1 model \cite{jiang2024mixtral} led with a 70\% initial step success and 45\% overall task completion, highlighting the importance of model size and domain knowledge for task-specific performance.
\begin{table}
\centering  
\caption{Ablation and Model Study on API Execution}  
\begin{tabular}{llcc}  
\toprule  
Type & Module & FS (\%) & OV (\%) \\  
\midrule  
\multirow{2}{*}{Ablation} & LLMs + UAR & 35 & 0 \\  
 & LLMs + UAR + RHR & \textbf{100} & \textbf{80} \\  
\midrule
\multirow{6}{*}{Models} & Qwen1.5B\cite{qwen} & 30 & 0 \\ 
 & Llama2-7B\cite{touvron2023llama} & 50 & 10 \\ 
 & Llama2-13B\cite{touvron2023llama} & 65 & 10 \\  
 & Mistral-7B-v0.1\cite{jiang2023mistral} & 65 & 20 \\ 
 & Mixtral-8x7B-Instruct-v0.1\cite{jiang2024mixtral} & 70 & 45 \\  
\bottomrule  
\end{tabular}  
\label{tab:combined_study}  
\end{table}
\section{Conclusion}

In this work, we propose a novel embodied intelligence system to improve ultrasound robotics by integrating the LLM and domain knowledge. First, our ultrasound operation knowledge database enables the LLM to perform precise motion planning, ensuring accurate and efficient scanning procedures. Then, our dynamic ultrasound scanning strategy empowers the LLM to dynamically adjust motion planning strategies during the scanning process, adapting to real-time observations and optimizing the scanning quality. Our system addresses the limitations of current ultrasound robots by enabling them to understand human intentions and instructions, thereby facilitating autonomous ultrasound scanning.


\bibliographystyle{IEEEtran}
\bibliography{ref}

\addtolength{\textheight}{-12cm}   






\end{document}